\def\year{2022}\relax
\documentclass[letterpaper]{article} % DO NOT CHANGE THIS
\usepackage{aaai22}  % DO NOT CHANGE THIS
\usepackage{times}  % DO NOT CHANGE THIS
\usepackage{helvet}  % DO NOT CHANGE THIS
\usepackage{courier}  % DO NOT CHANGE THIS
\usepackage[hyphens]{url}  % DO NOT CHANGE THIS
\usepackage{graphicx} % DO NOT CHANGE THIS
\usepackage{natbib}  % DO NOT CHANGE THIS AND DO NOT ADD ANY OPTIONS TO IT
\usepackage{caption} % DO NOT CHANGE THIS AND DO NOT ADD ANY OPTIONS TO IT
\DeclareCaptionStyle{ruled}{labelfont=normalfont,labelsep=colon,strut=off} % DO NOT CHANGE THIS
\usepackage{algorithm}
\usepackage{algorithmic}

\usepackage{amsmath}
\usepackage{latexsym}
\usepackage{booktabs}
\usepackage{url}
\usepackage{amsmath}

\usepackage{graphicx}

\usepackage{microtype}

\usepackage{graphicx}
\usepackage{amsmath}
\usepackage{amsfonts}
\usepackage{array}
\usepackage{comment}
\usepackage{url}
\usepackage{amsmath}
\usepackage{mathtools}

\usepackage{graphicx}
\usepackage{amsmath}
\usepackage{amsfonts}
\usepackage{array}
\usepackage{comment}

\usepackage{amsmath}
\usepackage{latexsym}
\usepackage{booktabs}
\usepackage{url}
\usepackage{amsmath}
\usepackage{hyperref}[2012/08/18]

\usepackage{newfloat}
\usepackage{listings}
\floatstyle{ruled}
\newfloat{listing}{tb}{lst}{}
\floatname{listing}{Listing}
\title{Tutorial Recommendation for Livestream Videos\\using Discourse-Level Consistency and Ontology-Based Filtering}
\author{
    Amir Pouran Ben Veyseh, \textsuperscript{\rm 1}\\
    Franck Dernoncourt\textsuperscript{\rm 2}, and Thien Huu Nguyen\textsuperscript{\rm 1}
}
\affiliations{
    \textsuperscript{\rm 1}Department of Computer and Information Science, University of Oregon\\

    \textsuperscript{\rm 2}Adobe Research\\

    apouranb@cs.uoregon.edu
}

\usepackage{bibentry}
\begin{document}

\maketitle

\begin{abstract}
Streaming videos is one of the methods for creators to share their creative works with their audience. In these videos,  the streamer share how they achieve their final objective by using various tools in one or several programs for creative projects. To this end, the steps required to achieve the final goal can be discussed. As such, these videos could provide substantial educational content that can be used to learn how to employ the tools used by the streamer. However, one of the drawbacks is that the streamer might not provide enough details for every step. Therefore, for the learners, it might be difficult to catch up with all the steps. In order to alleviate this issue, one solution is to link the streaming videos with the relevant tutorial available for the tools used in the streaming video. More specifically, a system can analyze the content of the live streaming video and recommend the most relevant tutorials. Since the existing document recommendation models cannot handle this situation, in this work, we present a novel dataset and model for the task of tutorial recommendation for live-streamed videos. We conduct extensive analyses on the proposed dataset and models, revealing the challenging nature of this task.
\end{abstract}

\section{Introduction}

\begin{figure*}
    \centering
    \includegraphics[scale=0.4]{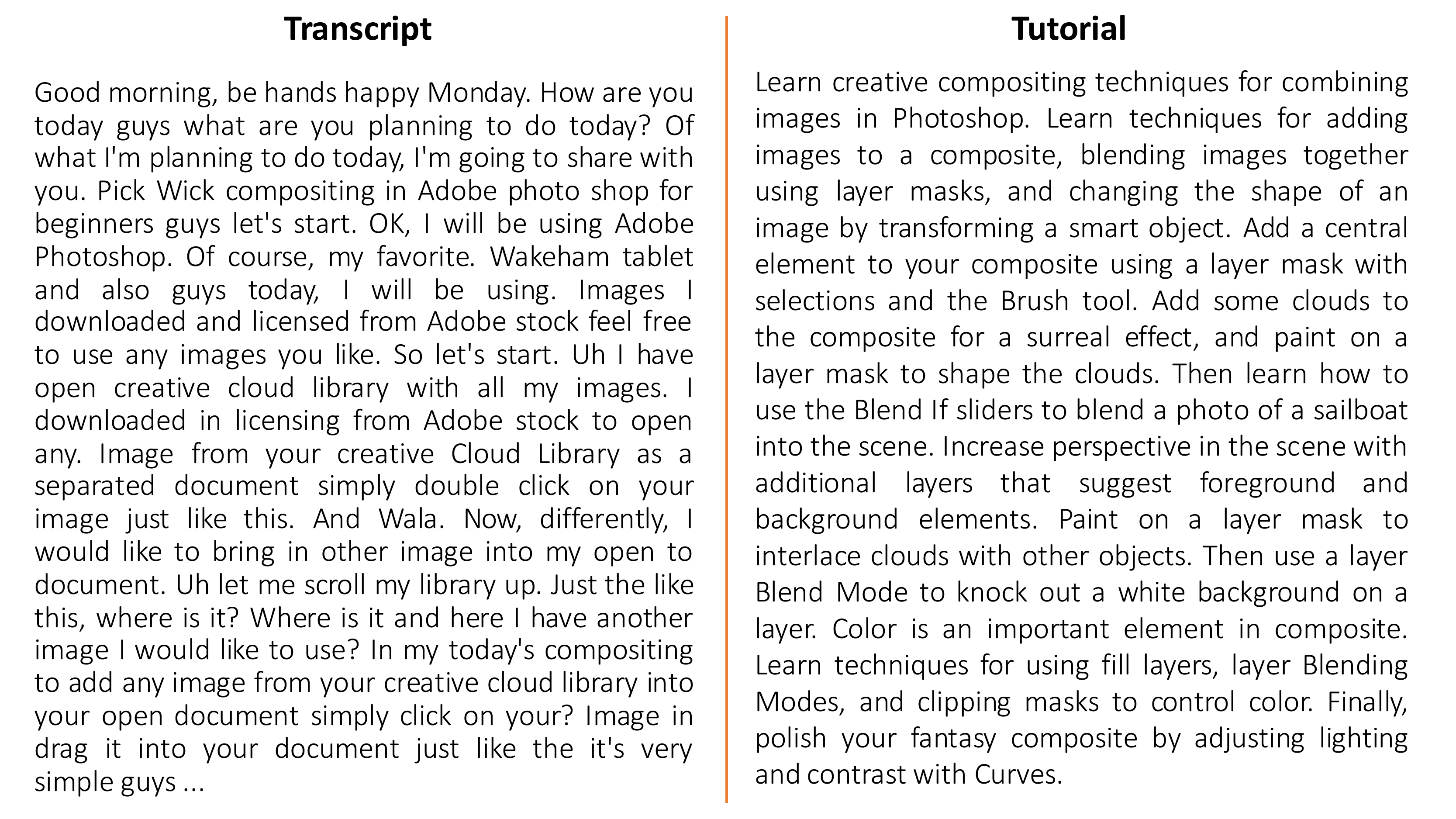}
    \caption{Part of the transcript of the live-streamed video "\href{https://www.behance.net/videos/1e245284-67e1-4d5c-a041-c7e0b6b72de4/Adobe-Photoshop-Compositing-for-Beginners}{Adobe Photoshop Compositing for Beginners}''
    % and the textual content of the tutorial ``\href{https://helpx.adobe.com/photoshop/how-to/compositing.html}{Compositing}".
    }
    \label{fig:example}
\end{figure*}

Streaming platforms, such as Twitch, Behance, and YouTube, are effective tools for creators that equip them with the facility of directly reaching out to their audience to share their creative content. For instance, on Behance\footnote{http://www.behance.net}, creators can share their works on visual projects, such as illustrations and designs, while employing visual content editing tools, such as Photoshop and Illustrator. In these videos, the streamer discusses the details of actions required to fulfill the objective of the creative task (e.g., designing a logo). Depending on the tools and the format that the streamer chooses to present their work, the streaming video can serve as educational content to learn how to use the tools used in the video. For instance, the streamer might review how to draw the sketches for designing a fantasy character or they might discuss the various methods for selecting an object in an image. Thus, these videos can help the audience to learn the nuances of the tools. However, the edit actions might be discussed in different details. For instance, to add some shapes to an image, the streamer might briefly mention the name of the brush employed to perform this action or he/she might explain the various methods available for this action. As such, a streaming video on itself might lack all details necessary to learn an edit action. One way to fill this gap is to accompany the streaming videos with tutorials in which the details of the actions are presented (see Figure \ref{fig:example}). Linking a streaming video with the relevant tutorials helps the audience to learn all aspects of the tools employed in the video.

Given a live streaming video, we aim to find the relevant tutorial to it. One solution to this question is to employ the existing document recommendation tools \cite{guan2010document,kim2016convolutional,xu2020understanding}. However, one limitation to this approach is that the existing recommendation tools are trained on formal documents, e.g., books or news articles. As such, directly employing these models for the recommendation for the transcripts of the live-streamed videos is not optimal. In particular, unlike formal texts, in the transcripts of a video, there might be incomplete sentences, incorrect words due to the ASR (automatic speech recognition) errors, or repeated sentences. These differences require domain-specific models that are designed to handle the challenges of the domain of video transcript. Moreover, another limitation for employing existing document recommendation resources is that there is no evaluation benchmark for this domain, making it more difficult to compare the performance of different models.

To address these shortcomings, in this work, we present the first large-scale tutorial linking dataset for the videos streamed on the Behance platform. More specifically, 47,403 sentences from the transcripts of 24 live-stream videos are annotated. In total, 4,126 sentences are annotated with 3 different tutorials for Photoshop. In addition, we also conduct extensive experiments on the proposed dataset. In particular, we first present the performance of an unsupervised model in which the similarity of the video transcript and the tutorial content is employed for the recommendation. Next, we employ the annotated data, to provide recommendations for the sentences in the video transcripts. Our analysis shows the challenging nature of this domain.

\section{Related work}

This task could be modeled either as text classification \cite{zhang2015character} or text similarity \cite{shahmirzadi2019text}. For text classification, the goal is to classify the input text into one of the pre-defined categories. Here, the categories might be defined as the available tutorials. The textual content of the tutorial describes the label of the category. For text similarity, the degree to which the tutorial content is similar to the video transcript is employed to find the most related tutorial for a given transcript. However, these solutions suffer from critical weaknesses which renders them inapplicable or inefficient for our task. First, most of the existing systems for text classification require manually labeled data. However, for our task, there is no human supervision available for training. As such, these methods might not be employed for this task. Second, both text classification and text similarity methods are evaluated on short documents (a few sentences) with formal language (i.e., a news article). However, in our task, the documents might be very large (e.g., transcripts of several hours of videos) and noisy (due to the automatic transcripts). This difference in a domain makes the majority of the solutions inapplicable to our task. Last but not least, the existing similarity-based methods cannot incorporate background knowledge (e.g., an ontology of concepts or keywords in the domain of the videos). Moreover, they ignore the discourse-level consistency between the two texts to compute the similarity score.

\begin{figure*}
    \centering
    \includegraphics[scale=0.4]{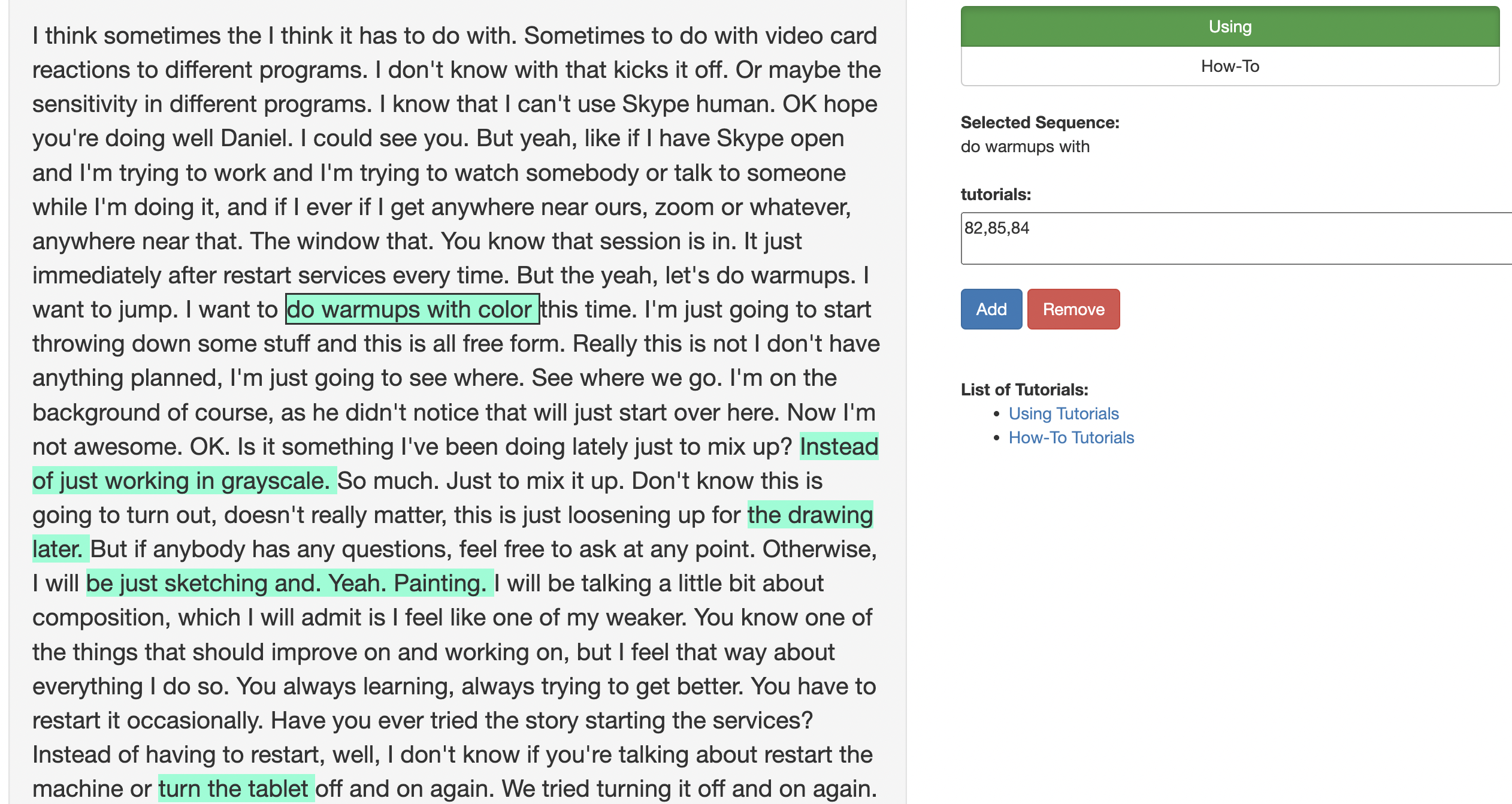}
    \caption{Annotation tool for phrase-level tutorial recommendation}
    \label{fig:tool}
\end{figure*}

\section{Data Annotation}

\subsection{Data Collection}

To train and evaluate the model we annotate data from the transcripts of the videos streamed on the Behance platform. The recordings are spilled by specialists and creators to share/discuss their inventive projects. As such, verbal substance from the speakers (in English) is imperative for video understanding. Whereas the recordings have introductory subjects, their substance is impromptu, thus the streamer might cut sentences, examine numerous themes, and utilize casual expressions. The recordings have an average length of 48 minutes. To get the verbal substance of the streamed recordings, we utilize the Microsoft ASR tool. In addition up to, 24 recordings (whose main editing tool is Photoshop) are transcribed. In total, 47,403 sentences are present in the transcribed videos to be annotated by human annotators.

\subsection{Data Annotation}

To annotate data, we hire expert annotators in Upwork who have experience in using Photoshop and also have experience in data annotation tasks. In total, three annotators are hired for this task. Every video transcript is assigned to the three annotators to link them to relevant tutorials. More specifically, for the tutorial pool, we employ two types of tutorials available for Photoshop:

\begin{itemize}
    \item Using: In this type of tutorial the usage of the specific tools is discussed. These tutorials are helpful to discuss the details of the tools that are used by the streamer. An example of this type of tutorial is presented in Figure \ref{fig:using_example}. In this work, we employ 290 Using tutorials.
    \item How-To: In this type of tutorial, the process to achieve a final edit action is discussed. For instance, how to design a portrait could be an edit action discussed in a tutorial. For this type of tutorial, multiple edit tools might be employed. An example of this tutorial is presented in Figure \ref{fig:howto_example}. In this work, we employ 126 How-To tutorials.
\end{itemize}

\begin{figure*}
    \centering
    \includegraphics[scale=0.4]{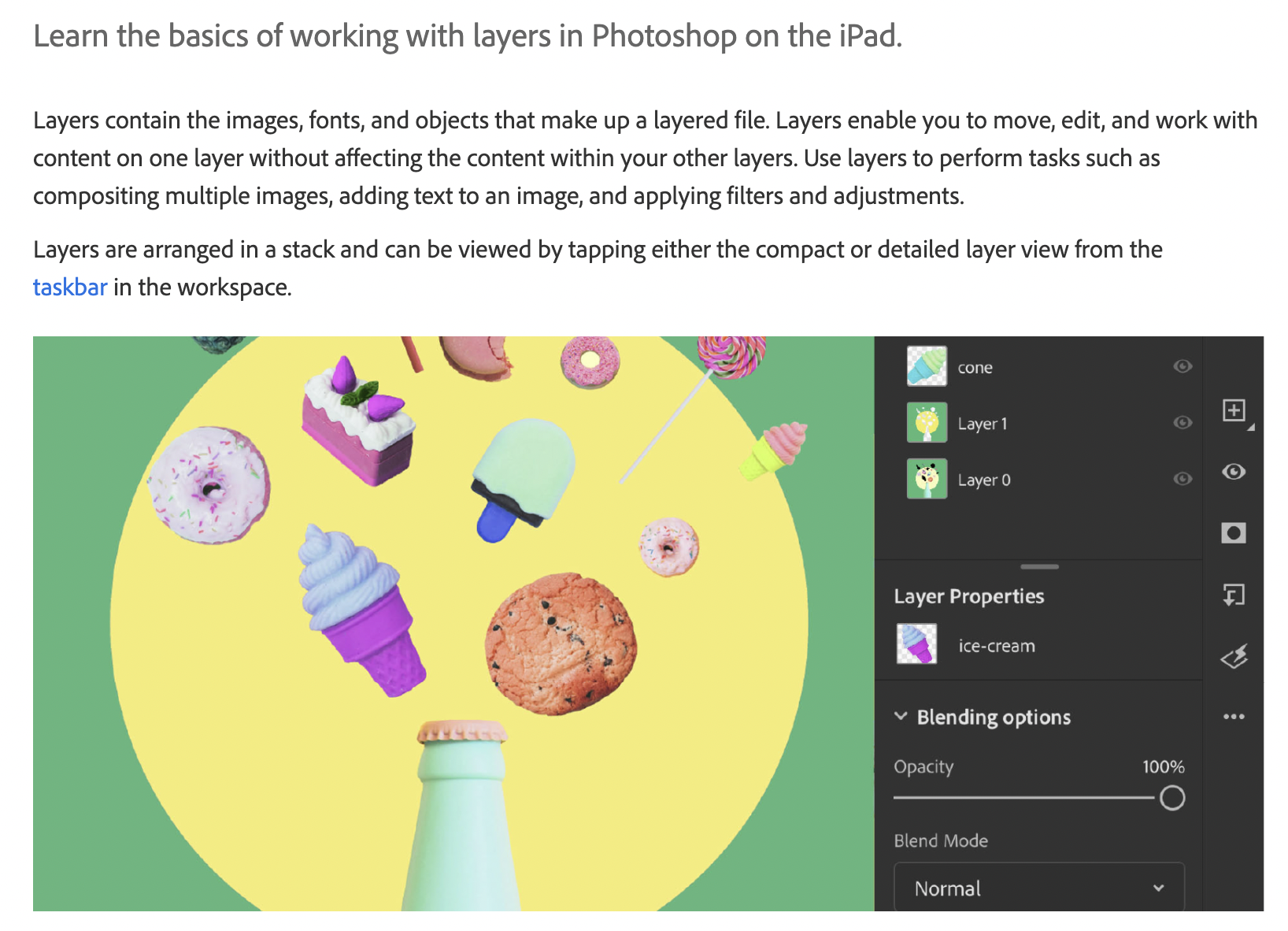}
    \caption{A Using tutorial. The tutorial discuss the application of layers.}
    \label{fig:my_label}
\end{figure*}

\begin{figure*}
    \centering
    \includegraphics[scale=0.4]{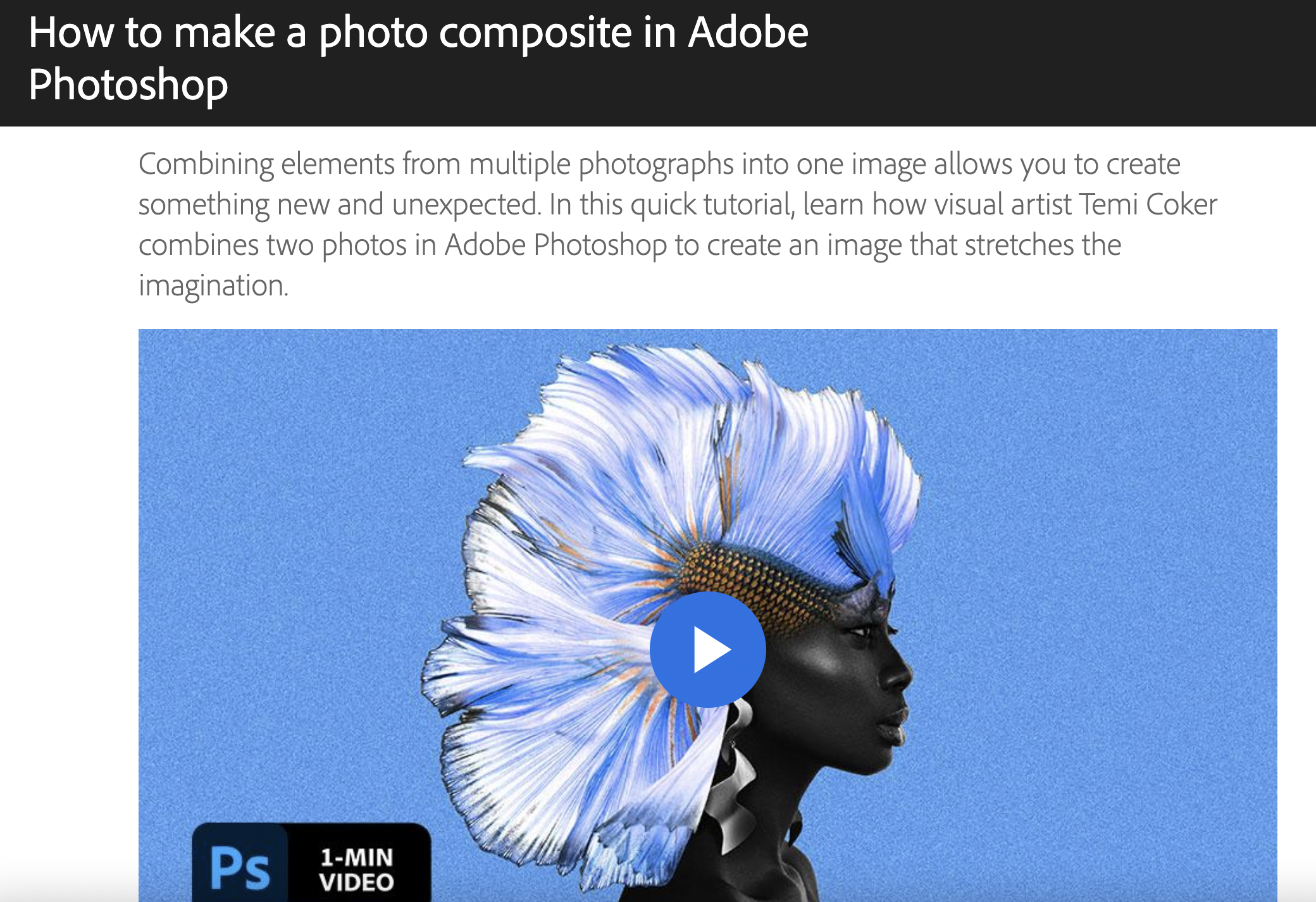}
    \caption{A How-To tutorial. The tutorial discuss the process of creating a composite image.}
    \label{fig:my_label}
\end{figure*}

For every transcript, the annotator selects the sentences that might refer to a tutorial and provide at least three tutorial related to the selected sentence. In total, 4,126 sentences with 3 tutorials are annotated.

\section{Model}

We employ two types of models: (1) An unsupervised model: In this model, the content of the transcript and the tutorial are employed for linking. This method provides a tutorial for the entire transcript; (2) A supervised model: In this model, we employ the annotated data to train a sentence classification model in which the model selects one of the available tutorials for a given sentence. If the sentence is not referring to a tutorial, $None$ is selected as the sentence label. The rest of this section provides details of these two models.

\subsection{Unsupervised Model Overview}

Our proposed model has the following novelties:

\begin{itemize}
    \item A novel unsupervised deep learning model for the task of tutorial recommendation using video transcripts
    \item A novel approach to employing domain-specific knowledge for filtering the target tutorials
    \item A novel method for summarizing the input transcript into a smaller version without using any human-curated labels
    \item A novel method for computing the similarity between the transcript and the tutorial text based on discourse-level consistency.
\end{itemize}

\subsection{Details}

Formally, the input to the system is the transcript $D=[w^D_1,w^D_2,\ldots,w^D_n]$, consisting of $n$ words, and a pool of tutorial textual content, i.e., $P=[T_1,T_2,\ldots,T_{|P|}]$ where $T_i=[w^T_1,w^T_2,\ldots,w^T_m]$ is the textual content of $i$-th tutorial consisting of $m$ words. The goal is to return the most relevant tutorial from the pool $P$, i.e., $T_{gold}$. To create a system for this task, we propose multiple components. Specifically, the proposed system consists of three major components:

\begin{itemize}
    \item Filtering Tutorials: In this component, the goal is to remove the tutorials that are very unlikely to be relevant to a given transcript. In this component, domain-specific knowledge is employed to assess the relevancy between the given transcript and the tutorial textual content.
    \item Transcript summarizing: In this component the objective is to summarize the given transcript such that only the most important information that could be helpful for finding the tutorials are preserved. This component employs an unsupervised deep learning method.
    \item Tutorial Ranking: Finally, using the summary of the given transcript and the filtered list of relevant tutorials, this component employs various metrics to sort the tutorials based on their relevance and similarity to the transcript.
\end{itemize}

The rest of this section elaborates more on the details of each of these components.

\subsubsection{Filtering Tutorials}

The pool of tutorials might contain several irrelevant candidates which could hinder an efficient ranking method. As such, it is necessary to first filter the pool $P$ such that the unlikely candidates are removed. Formally, the objective is to define the filter function $\Phi$ such that:

\begin{equation}
    \begin{split}
        |\Phi(P)| &< |P| \\
        \forall \text{ } T_i \in \Phi(P)\text{, } T_j \in P-\Phi(P) & \quad R(T_i) < R(T_j)
    \end{split}
\end{equation}
where $|\cdot|$ is the size of the pool and $R(x)$ is the rank of $x$ in the sorted pool based on the relevance of the candidates to the given transcript $D$. To define the function $\Phi$, in a novel method, we propose to employ two types of criteria:

\textbf{Domain-specific Knowledge}: The ontology of names in the domains of interest (i.e., in our work we use Image Editing Tools such as Adobe Photoshop as the domain of interest) could be employed as the domain-specific knowledge. Specifically, in this work, we propose to employ the available list of tool names for Adobe Photoshop as domain-specific information. To employ this knowledge in function $\Phi$, we suggest first find the tool names mentioned in the transcript $D$, i.e., $TN=\{tn_1,tn_2,\ldots,tn_{|TN|}\}$ where $tn_i$ is a tool name in transcript $D$. Next, we define the filter function $\Phi_{DK}$ to filter out those tutorials in the pool $P$ which don't mention one of the tool names of $TN$:

\begin{equation}
    \Phi_{DK}(T_i) = \left\{
    \begin{array}{ll}
        True & \text{if } \exists \text{ } tn_j \in TN \text{ s.t. } tn_j \in T_i \\
        False & \text{otherwise}
    \end{array} \right.
\end{equation}

\textbf{String Similarity}: In addition to the domain-specific knowledge, we seek to incorporate string similarity of the tutorial textual content with the transcript $D$ in the filtering process. More specifically, we first compute the string similarity of the transcript $D$ with tutorial $T_i \in P$ using normalized point-wise mutual information (PMI):

\begin{equation}
    \begin{split}
        Sim(D,T_i) & = \sum_{w^D_i \in D} \sum_{w^T_j \in T_i} \frac{PMI(W^D_i,W^T_j)}{n*m} \\
        PMI(w_i,w_j) & = \log \frac{COUNT(w_i,w_j)}{COUNT(w_i)*COUNT(w_j)}
    \end{split}
\end{equation}
where $n$ and $m$ are the number of words in the transcript $D$ and tutorial $T_i$, $COUNT(W_i)$ is the number of occurrences of the word $w_i$ in all transcripts in the training data and $COUNT(w_i,w_j)$ is the number of occurrences of both words $w_i$ and $w_j$ in a transcript in the training data. Next, we define the following filter function based on similarity:

\begin{equation}
    \Phi_{sim}(T_i) = \left \{
    \begin{array}{ll}
        True & \text{if } Sim(D,T_i) > \delta \\
        False & \text{otherwise}
    \end{array}
    \right.
\end{equation}
where $\delta$ is a hyper-parameter to be tuned using development data. Finally, using the two aforementioned filter functions $\Phi_{DK}$ and $\Phi_{sim}$, we define the final function $\Phi$:

\begin{equation}
    \Phi(T_i) = \left \{
    \begin{array}{ll}
        True & \text{if } \Phi_{DK}(T_i) \text{ and } \Phi_{sim}(T_i)\\
        False & \text{otherwise}
    \end{array}
    \right.
\end{equation}

\subsubsection{Transcript Summarizer}

The next component in our system is the transcript summarizer whose goal is to shorten the transcript $D$ such that only the distinctive information is preserved and the redundant portions, which might not be helpful to identify the most related tutorial, are excluded. To this end, we proposed to train a deep learning model to consume the input document $D$ and generate the shorter document $D'$, such that $|D'| < |D|$. Specifically, a transformer-based language model, i.e., BERT, is trained to encode the words of the document $D$, i.e., $e_i = BERT(w_i)$ for all $w_in \in D$\footnote{Note that for words consisting of multiple word-pieces, we represent them using the average of their word-piece embeddings obtained from BERT model}. In our experiments, we use the last hidden states of the BERT model to obtain the embedding vector $e_i$. Next, a feed-forward network is utilized to estimate the likelihood of the word $w_i$ to be included in the shorter document $D'$:

\begin{equation}
    P(w_i|D) = FF(w_i) = \sigma(W_1*(W_2*e_i+b_2)+b_1)
\end{equation}
where $\sigma$ is the sigmoid activation function, $W_1$ and $W_2$ are the weight matrices and $b_1$ and $b_2$ are bias. To train the BERT model and the feed-forward layer, since the available resources for this task do not have any labeled data, we resort to unsupervised learning. Concretely, two criteria are employed to train the model:

\textbf{Distinctiveness}: The shorter documents $D'_i$ and $D'_j$ obtained from the original documents $D_i$ and $D_j$ should be as different as possible. Moreover, those portions of the documents $D_i$ and $D_j$ that do not appear in the summary should have the least differences (in other words, as these portions are not informative (e.g., they are chitchat), there will be more similarity between them). To fulfill this requirement, we employ the following loss function:

\begin{equation}
    \mathcal{L}_{dist} = \alpha * ( \sigma(H'_i) \odot \sigma(H'_j) ) - \beta (\sigma(H''_i) \odot \sigma(H''_j))
\end{equation}
where $\sigma$ is the softmax function, $\alpha$ and $\beta$ are the trade-off parameters and $\odot$ is the hadamard product. The vectors $H'_i$, $H'_j$, $H''_i$ and $H''_j$ are the vector representation for the summaries $D'_i$, $D'_j$, and the portions of the documents $D_i$ and $D_j$ not included in the summaries, i.e., $D''_i$ and $D''_j$, respectively. To obtain these vector representations, we use max-pooling on the multiplication of the embeeding vectors and the feed-forward network. For instance, $H'_i$ is obtained as follows:

\begin{equation}
    \label{eq:vec}
    \begin{split}
        H'_i &= MAX\_POOL(h_1,h_2,\ldots,h_n) \\
        h_k & = e_k * FF(e_k)
    \end{split}
\end{equation}
where $n$ is the number of words in the document $D$. Note that for the representations of $H''_i$ and $H''_j$, we replace $FF(e_k)$ with $(1-FF(e_k))$ in equation \ref{eq:vec}.

\textbf{Information Retaining}: The process to summarize document $D$ into the smaller version $D'$ is supposed to keep the most important information in $D$ intact. As such, it is expected that the information available in both documents $D$ and $D'$ to have considerable overlap. This criterion can be achieved via increasing the mutual information (MI) between the representations of the $D$ and $D'$. To fulfill this goal, we exploit contrastive learning. In particular, a discriminator is trained to distinguish positive samples from negative ones where a positive sample is the concatenation of the representation of the original document $D_i$ and its summary $D'_i$, i.e., $pos=[H_i:H'_i]$, and the negative sample is the concatenation of the representation of the document $D_i$ with the summary of the randomly selected document $D_j$, i.e., $neg=[H_i:H'_j]$. Formally, the following loss function is employed to increase the mutual information:

\begin{equation}
    \label{eq:cont}
    \mathcal{L}_{IR} = -(log(\Psi[H_i:H'_i])+log(1-\Psi([H_i:H'_j])))
\end{equation}
where $\Psi$ is the discriminator. The sum of two losses, i.e., $\mathcal{L}_{dist}$ and $\mathcal{L}_{IR}$ is employed as the final loss function to train the transcript summarizer component:

\begin{equation}
    \mathcal{L} = \alpha \mathcal{L}_{IR} + \beta \mathcal{L}_{dist}
\end{equation}
where $\alpha$ and $\beta$ are trade-off parameters. At inference time, to summarize the input transcript, we employ the output of the feed-forward network $FF$ and every word $w_i$ whose corresponding value from $FF$ is higher than a threshold is selected in $D'$.

\subsubsection{Ranking}

Finally, given the short document $D'$ and the filtered list of tutorials, i.e., $P' = \Phi(P)$, the final component aims to sort the tutorials based on their relevance to the given transcript. To this end, we employ two types of scores:

\begin{itemize}
    \item String Similarity: The string similarity between every tutorial $T_i \in P'$ with the summary $D'$ is evaluated using fasttext \cite{bojanowski2016enriching} to obtain its string similarity score: $Score_{str} = FastText(T_i,D')$
    \item Discourse Similarity: The objective of this score is to measure how likely is the tutorial $T_i \in P'$ to complement the summary $D'$ of the input transcript. To this end, first, we train the text classification model $C$ that takes the concatenation of the first and second half of the transcripts $D_i$, i.e., $D_{i,1}$ and $D_{i,2}$, as a positive sample and the concatenation of $D_{i,1}$ and $D_{j,2}$, where $j$ is selected randomly, as the negative samples\footnote{words of the documents are encoded using GloVe embedding and the max-pooled embeddings of them are fed into classifier $C$}. The model is trained using a similar loss function as equation \ref{eq:cont}. Next, the trained classifier $C$ is employed in the ranking component by feeding the concatenation of the tutorial $T_i$ and summary $D'$ to the model and its output (i.e., the likelihood of the input to complement each other), is employed as the discourse-level score: $Score_{disc} = C([T_i:D'])$
\end{itemize}

Finally, to sort the documents, we compute the sum of the two aforementioned scores:

\begin{equation}
    Score = Score_{sim} + Score_{disc}
\end{equation}

The sorted list of the tutorials is returned as the final output of the system.

\subsection{Supervised Model Overview}

To train the model for supervised tutorial linking, we model the task as a sentence classification problem. More specifically, given the words of the sentence $S$ and the title of the tutorial $T$, the input sequence $[CLS],w^S_1,w^S_2,\ldots,w^S_n,[SEP],w^T_1,w^T_2,\ldots,w^T_m,[SEP]$ is fed into the BERT$_{base}$ model. Note that for every tutorial in the tutorial pool a separate input is prepared. Next, the $[CLS]$ vector representation is fed into a two-layer feed-forward layer to predict a binary label, e.g., 1/0. The label is 1 if the tutorial is linked with the given sentence in the annotated dataset. Note that we add a special tutorial title $None$ for sentences without any linked tutorial.

\subsection{Experiments}

\begin{table}[]
    \centering
    \begin{tabular}{l|c|c}
       Model & Hit@3  & Hit@5 \\ \hline
       String Similarity Sorting  & 40\%  & 50\% \\
       Keyword Sorting & 35\% & 45\% \\
       Information-based Sorting & 40\% & 50\% \\ \hline
       Ours & 55\% & 65\%
    \end{tabular}
    \caption{Performance of the unsupervised models}
    \label{tab:results}
\end{table}

\begin{table}[]
    \centering
    \begin{tabular}{l|c}
       Model & F1 \\ \hline
       BERT  & 35\% \\
       GloVe & 28\% \\
    \end{tabular}
    \caption{Performance of the supervised models}
    \label{tab:results-supervised}
\end{table}

To evaluate the proposed system, we manually annotated transcripts from the Behance corpus. These are the transcripts of videos streamed on Behance.net and the streamers are all using Adobe Photoshop in their streaming video. We use Adobe Photoshop tutorials (more than 200 tutorials for using or how-to) as the initial pool of tutorials. To provide more insight into the performance of the proposed system, we compare it with the following systems:

\begin{itemize}
    \item String Similarity Sorting: In this system, the string similarity of the input transcript and the tutorials are measured and it is employed to sort all tutorials in the pool.
    \item Keyword Sorting: In this system, the tutorials are sorted based on the number of tool names that they have in common with the input transcript.
    \item Information-based Sorting: In this system, the same PMI-based scoring that is employed in our filtering component is employed to sort all tutorials.
\end{itemize}

For the supervised model, in addition to the proposed BERT model, we also compare the performance of the model when the words of the sentence $S$ and the title of the tutorial $T$ are presented by GloVe embedding. The max-pooled representation of the input is fed into the feed-forward network.

To evaluate the models, we use Hit@3 and Hit@5 evaluation metrics. % (e.g., Hit@3 is the number of cases that the most related tutorial is returned as amongst the top 3 most related tutorials in the sorted list of tutorials.)
The results are shown in Table \ref{tab:results}. This table clearly shows that the proposed model significantly outperforms the systems, indicating its effectiveness for this task.

The results for the supervised model are presented in Table \ref{tab:results-supervised}. This table shows that the contextualized representation of the video transcript and the title are more effective than the GloVe embedding. Nonetheless, both models suffer from low performance which indicates more research is required.

\bibliography{aaai22}

 \clearpage
 \clearpage
\newpage

\section{Appendices}

\section{Tutorials}

\begin{figure*}
    \centering
    \includegraphics[scale=0.4]{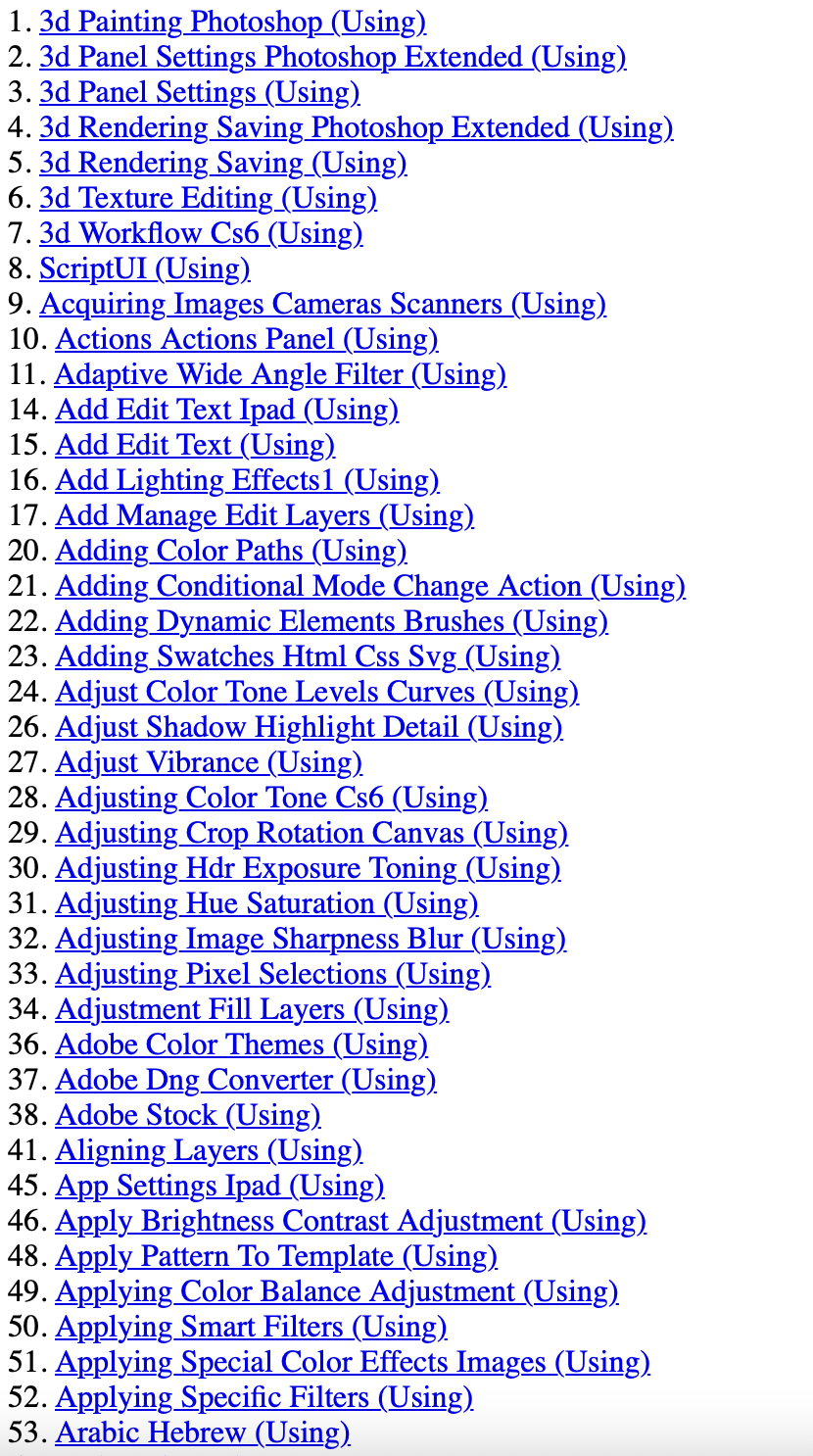}
    \caption{A sample list of Using tutorials}
    \label{fig:using}
\end{figure*}

\begin{figure*}
    \centering
    \includegraphics[scale=0.4]{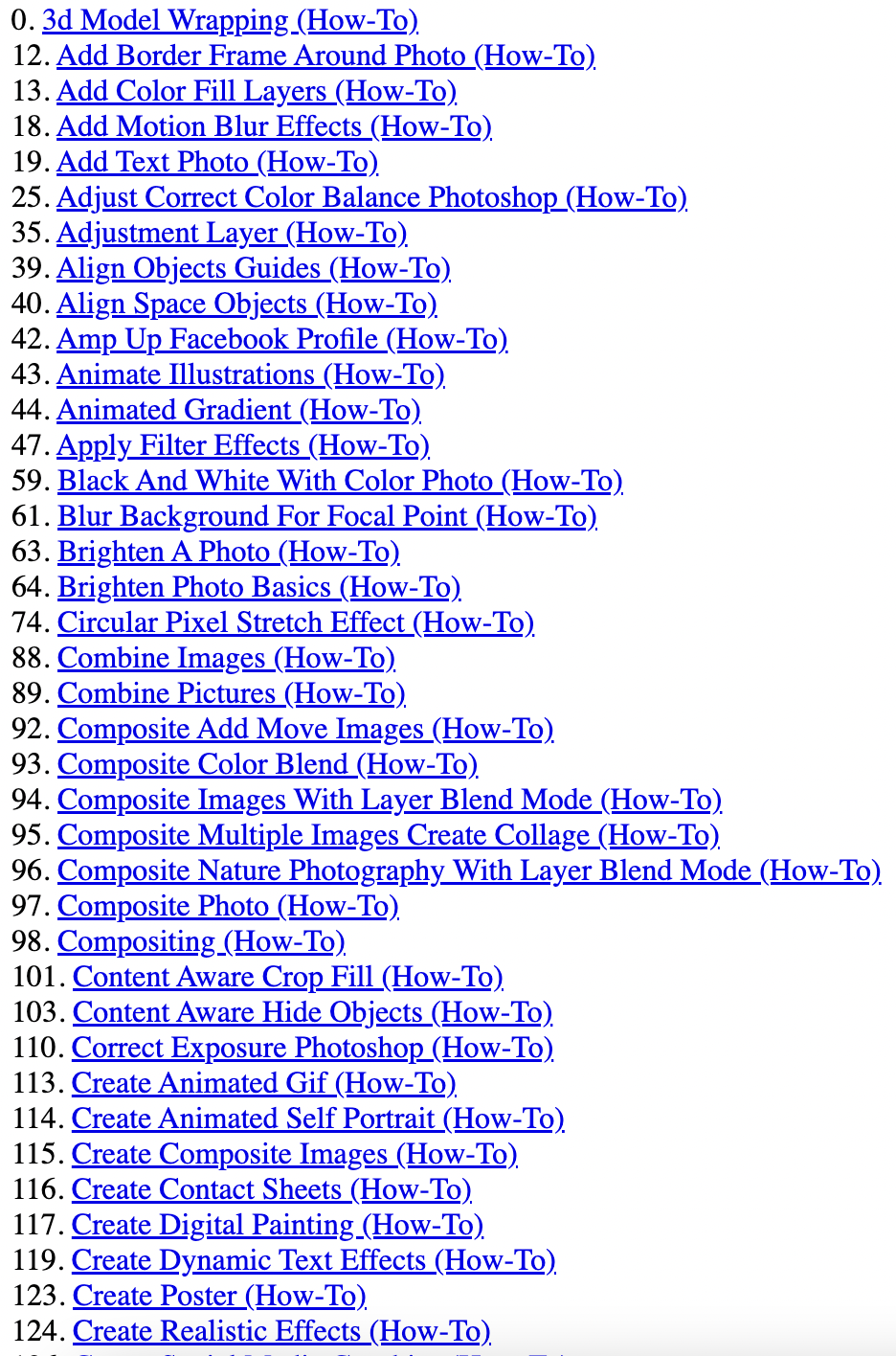}
    \caption{A sample list of HowTo tutorials}
    \label{fig:howto}
\end{figure*}

A list of Using and How-To tutorials employed during the annotation are shown in Figures \ref{fig:using} and \ref{fig:howto}. In our work, in total, 290 and 126 Using and How-To tutorials are employed, respectively.

\section{Case Study}

\begin{figure*}
    \centering
    \includegraphics[scale=0.4]{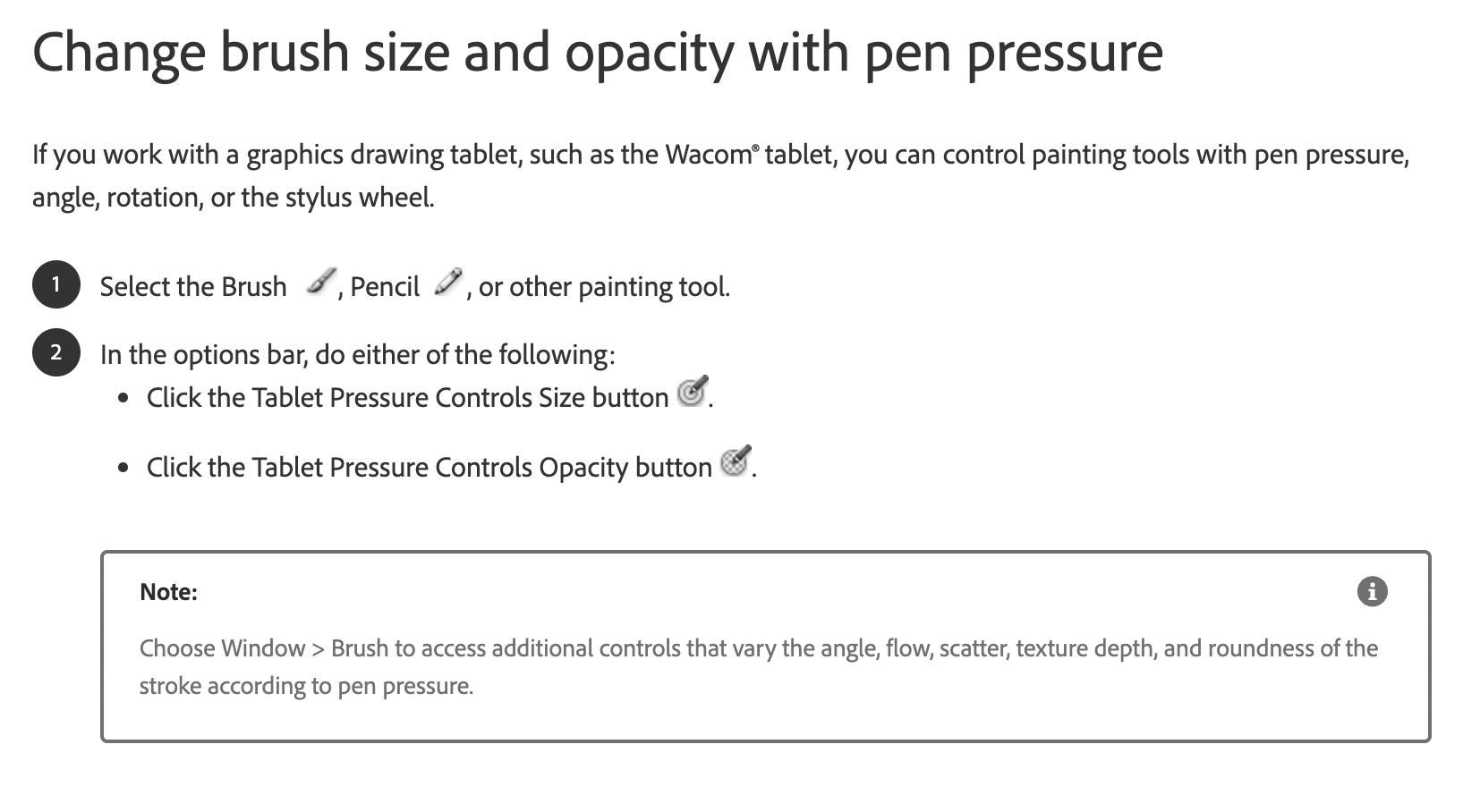}
    \caption{A screenshot of the tutorial page for the sentence shown in Case Study section.}
    \label{fig:tut1}
\end{figure*}

\begin{figure*}
    \centering
    \includegraphics[scale=0.4]{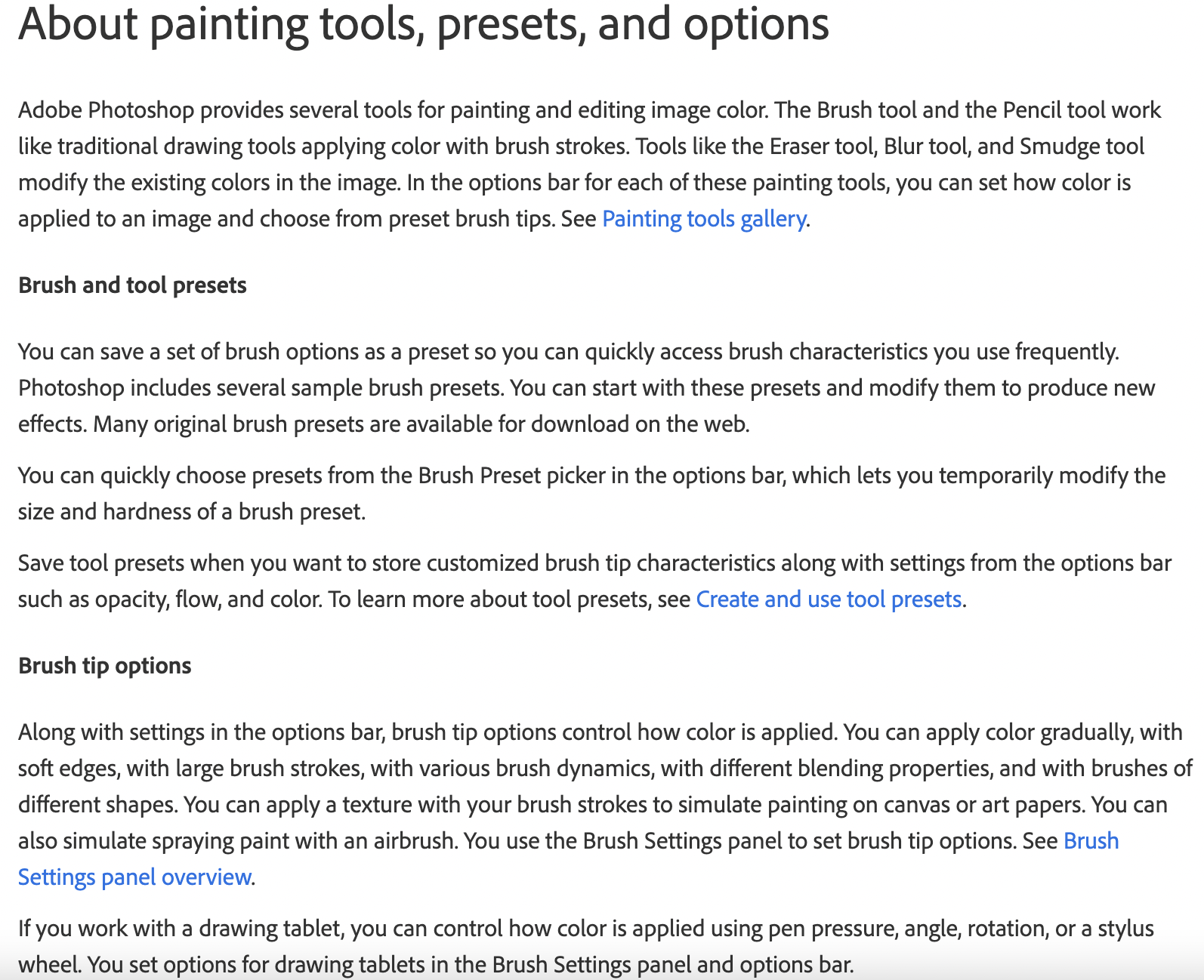}
    \caption{A screenshot of the tutorial page for the sentence shown in Case Study section.}
    \label{fig:tut2}
\end{figure*}

\begin{figure*}
    \centering
    \includegraphics[scale=0.4]{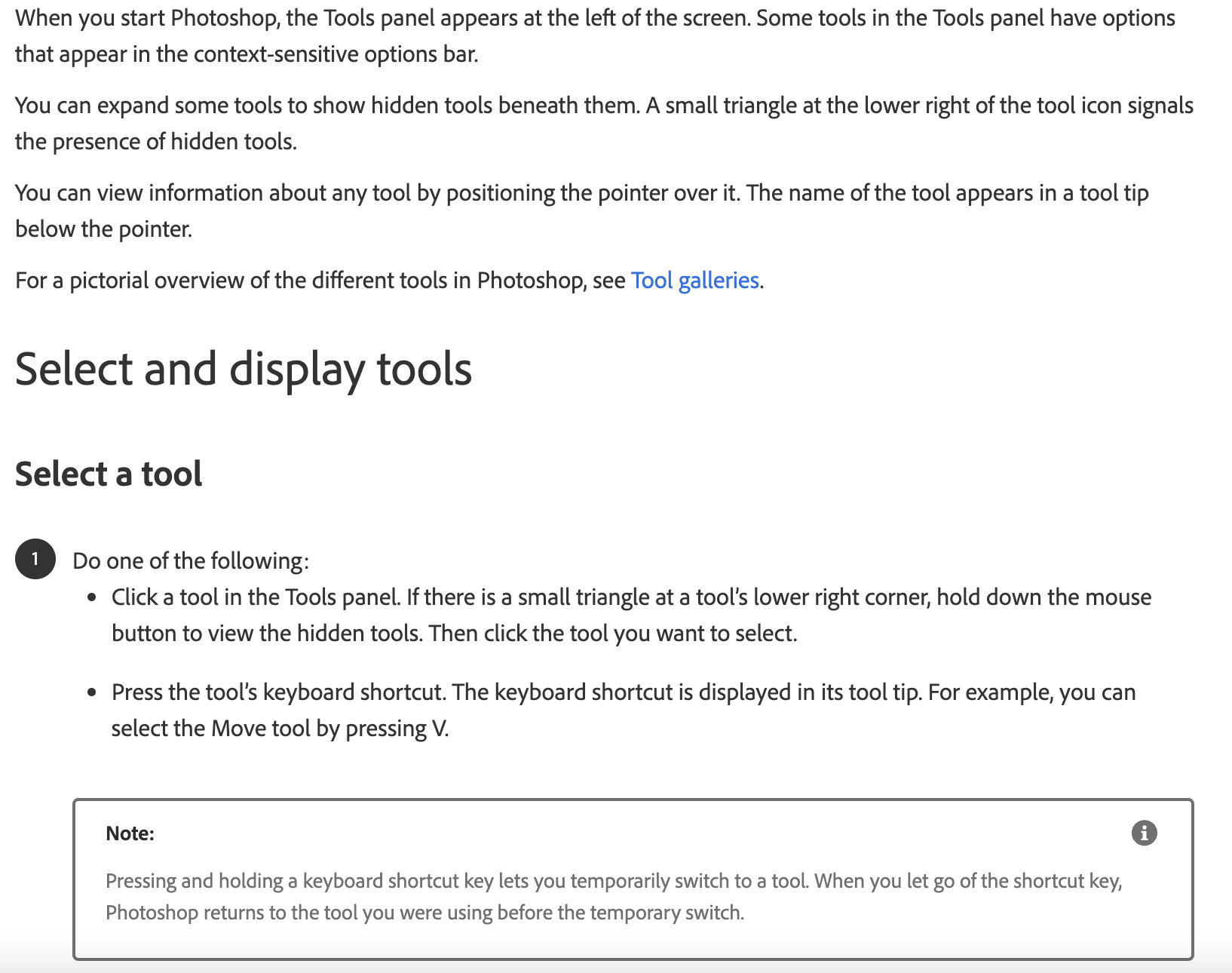}
    \caption{A screenshot of the tutorial page for the sentence shown in Case Study section.}
    \label{fig:tut3}
\end{figure*}

In order to provide more insight into the performance of the proposed supervised model, we show the predicted tutorials for the sentence ``\textit{I'm gonna work the composition of this and then work on cleaning up the drawing a little bit}" in the paragraph ``\textit{Right there. I get it. OK, I get it. I will do this real quick as the last thing. Change this brush. Might get a little bit right there. Something like that. Anyway. I don't know. So this is what we did for warmups. This guy in this lady. Will leave it at that. But now we will jump back into ours. A party planning here. They actually took a little hit longer than I want to be out about an hour and a half hour and 1520 minutes left. \textbf{I'm gonna work on the composition of this and then work on cleaning up the drawing a little bit}. Will pull this up. I think I still have it. No, it's not in here. Just to kind of recap. Z. Well, let's see. Let's see if I can find this cleaning up the drawing }". The results are shown in Figure \ref{fig:tut1}, \ref{fig:tut2} and \ref{fig:tut3}.

% \begin{figure*}
%     \centering
%     \includegraphics[scale=0.4]{sample.png}
%     \caption{A sample text from the video transcript. The highlighted phrase is annotated with three Using tutorials, shown in Figure \ref{fig:tut1,fig:tut2,fig:tut3}}
%     \label{fig:sampletext}
% \end{figure*}

\end{document}